\documentclass{article}

\usepackage{times}
\usepackage{amsmath}
\usepackage{graphicx} % more modern
\usepackage{subfigure} 

\usepackage{tabularx} 

\usepackage{natbib}

\usepackage{algorithm}
\usepackage{algorithmic}
\usepackage[algo2e]{algorithm2e} 
\usepackage{threeparttable}
\usepackage{booktabs}
\usepackage{wrapfig}
\usepackage{enumitem}

\providecommand{\SetAlgoLined}{\SetLine}

\usepackage[draft]{hyperref}
\usepackage[accepted]{icml2016} 

\icmltitlerunning{Modeling the Adoption of Financial Services for the Poor}

\begin{document} 

\twocolumn[
\icmltitle{Machine Learning Across Cultures:\\
Modeling the Adoption of Financial Services for the Poor}

\icmlauthor{Muhammad Raza Khan}{mraza@uw.edu}
\icmladdress{University of Washington,
            Seattle, WA 98195 USA}
\icmlauthor{Joshua E Blumenstock}{joshblum@uw.edu}
\icmladdress{University of Washington,
	Seattle, WA 98195 USA}

\icmlkeywords{Mobile Money; Feature engineering; Product adoption; Supervised learning, CDR}

\vskip 0.3in
]

\begin{abstract} 
Recently, mobile operators in many developing economies have launched ``Mobile Money'' platforms that deliver basic financial services over the mobile phone network. While many believe that these services can improve the lives of the poor, a consistent difficulty has been identifying individuals most likely to benefit from access to the new technology. Here, we combine terabyte-scale data from three different mobile phone operators from Ghana, Pakistan, and Zambia, to better understand the behavioral determinants of mobile money adoption. Our supervised learning models provide insight into the best predictors of adoption in three very distinct cultures. We find that models fit on one population fail to generalize to another, and in general are highly context-dependent. These findings highlight the need for a nuanced approach to understanding the role and potential of financial services for the poor.%\footnote{This workshop paper summarizes results in a longer paper recently submitted to KDD.}
% Over the last decade mobile  operators, developing agencies and government organizations have been trying to promote Mobile Money services as a cost-effective and efficient alternative to the traditional financial services. However, the Mobile Money Services have seen varying levels of adoption in countries with similar patterns of development. In this work, we have tried to develop a model to predict adoption of Mobile Money using a novel Deterministic Finite Automaton based algorithm for features engineering. Based on the results of our model on billions of call records from three different developing countries, we have demonstrated that our adoption model outshines the current established practices of product promotion and adoption. However, we do also find that the product adoption models do not generalize across different contexts. 
\end{abstract} 

%%%%%%%%%%%%%%%%%%%%%%%%%%%%%%%%
% INTRODUCTION
%%%%%%%%%%%%%%%%%%%%%%%%%%%%%%%%

%\vspace{.2in}

\section{Introduction}
\label{sec:intro}
%needs to be about 600 words

Billions of people around the world live without access to formal financial institutions. Over the last several years, mobile phone operators across the globe have launched ``Mobile Money" platforms, which make it possible for many of the world's poor to conduct basic financial transactions from inexpensive feature phones. However, outside of Kenya and a few other countries, worldwide adoption of Mobile Money has been extremely anemic. The vast majority of deployments have struggled to promote sustained product adoption, and an industry report from 2014 estimates that 66\% of registered customers were inactive \cite{scharwatt_state_2014}. An open and important question thus revolves around understanding what drives customers to adopt and use Mobile Money, and whether patterns observed in one country will generalize to another.

Our work builds on several distinct strands in the academic literature. The first strand that is concerned with understanding the determinants of mobile money adoption has historically been the domain of development researchers, and includes both macroeconomic studies \cite{mas_scaling_2011,dermish_branchless_2011,donovan_mobile_2012} as well as qualitative, ethnographic work \cite{morawczynski_exploring_2009,mobile_banking_adoption_mehdi,etim2014mobile}.
A second strand of literature seeks to derive general insights from mobile phone transactions logs. This encompasses a wide array of applications, including predicting the socioeconomic status \cite{blumenstock_predicting_2015}, gender \cite{frias2010gender}, and age \cite{dong_who_2014} of individual mobile phone users. 

Most relevant to this study, a third area of prior work mines digital transactions logs to model the social and behavioral determinants of product adoption \cite{ugander_pnas_2014,leskovec_dynamics_2007}, with a few recent papers written in the context of mobile phone data \cite{khan2015behavioral,sundsoy_big_2014,cgap}. Relative to these studies, our study moves this literature forward by (a) innovating in the method used to generate features, thereby providing a systematic and comprehensive approach to feature engineering; (b) leveraging data from three different contexts to calibrate the external validity and generalizability of our results; and (c) carefully articulating the experiments in a way that will enable other researchers to replicate and extend these methods.

%\begin{figure}[tb]%
%	\centering
%	\includegraphics[width=3.0in,height=1.6in]{./figs/world2.png}%
%	\caption{Worldwide access to formal financial services, constructed using data from the Global Findex Dataset \cite{demirgucc2015global_inclusion}. Study locations are identified by pins.}% Following part of the caption should be normal rather than bold %
%	\label{fig:worldmap}%
%\end{figure}

% \section{Related Work}
\label{sec:related}

%%%%%%%%%%%%%%%%%%%%%%%%%%%%%%%%
% DATA
%%%%%%%%%%%%%%%%%%%%%%%%%%%%%%%%
\section{Data and Context}
\label{sec:data}
The data that we have used in this project consists of anonymized Call Detail Records (CDR) and Mobile Money Transaction Records (MMTR) of all the subscribers from three different operators in Ghana, Pakistan, and Zambia. All three countries rank in the bottom third of the Human Development Index and Financial Inclusion Index as shown in Table \ref{tab:sumstats}. The CDR and MMTR contain basic metadata on every event that occurs on the mobile phone network, including phone calls, text messages, and any form of Mobile Money activity. In total, the original data contains billions of transactions conducted by tens of millions of unique individuals.  Each dataset spans several months of activity, which we divide into a ``training'' period and an ``evaluation'' period. CDR from a 10-day training period was used to engineer features and fit a predictive model, where the target variables (based on Mobile Money activity) were measured in a subsequent 3-month evaluation period. Based on activity during the evaluation period, each subscribers is categorized as a ``Voice Only User'' (no MM activity), a ``Registered Mobile Money User'' (one or more MMTR's), or an ``Active Mobile Money User'' (at least one MMTR in each month).

%\begin{figure}[htb]%
%	\centering
%	\includegraphics[width=3.3in]{./figs/gantt.png}%
%	\caption{Training and evaluation periods}%
%	\label{fig:gantt}%
%\end{figure}

%%%%%%%%%%% TABLE %%%%%%%%%%%%%

\begin{table}[htb!]
\begin{threeparttable}
\renewcommand{\arraystretch}{1.1}
\footnotesize
\resizebox{\columnwidth}{!}{%
\begin{tabular}{l*{4}{l}}
\addlinespace
\toprule
		\textbf{Country} & \textbf{Ghana} & \textbf{Pakistan} & \textbf{Zambia} \\
\midrule
\multicolumn{4}{l}{\textit{Panel A: National statistics (Source: World Bank)}}\\
Population & 25.90 Million & 185.00 Million & 15.72 Million\\
GDP PPP adjusted  & \$4081.70 & \$4811.4 & \$3904.00 \\
Mobile accounts per 100 users  & 115 & 73 & 67 \\
Adults without financial services & 59.49\%& 54.36\% & 86.97\%\\
%http://data.worldbank.org/indicator/IT.CEL.SETS.P2
\midrule
\multicolumn{4}{l}{\textit{Panel B: Mobile phone use (Source: Call Detail Records)}}\\
Calls per user per day & 6.53 (6.99) & 7.76 (10.25) & 10.26 (102.86) \\
SMS per user per day &3.10 (100.36) &38.71 (80.83) &10.88 (262.81) \\
No. of unique contacts  &21.66 (24.91) & 46.93 (139.67) & 17.63 (328.63) \\
No. of unique towers  & 12.98 (16.07) &24.15 (57.30) &7.35 (17.56) \\[0.1em]
\bottomrule
\end{tabular}
}
\begin{tablenotes}[normal,flushleft]
\scriptsize
\item \emph{Notes}: Standard deviations reported in parenthesis. 
\end{tablenotes}
\caption{Summary statistics by country\label{tab:sumstats}}
\end{threeparttable}
\end{table}
%%%%%%%%%%% TABLE %%%%%%%%%%%%%

%%%%%%%%%%%%%%%%%%%%%%%%%%%%%%%%
%% Feature engineering
%%%%%%%%%%%%%%%%%%%%%%%%%%%%%%%%%
\section{Feature Engineering}
\label{sec:features}
\begin{figure*}[htb!]
	\centering
	\begin{minipage}[t]{0.55\linewidth}
		\includegraphics[width=3.5in]{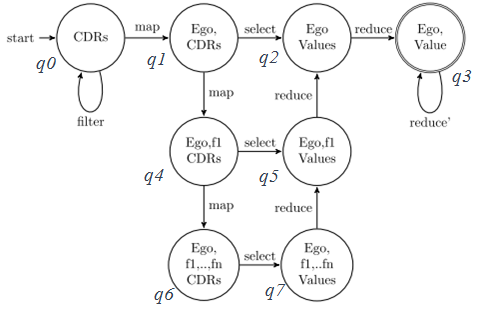}
		\caption{Deterministic Finite Automaton\label{fig:states}}	
	\end{minipage}
	%	\quad 
	\begin{minipage}[t]{0.4\linewidth}
		\includegraphics[width=3in]{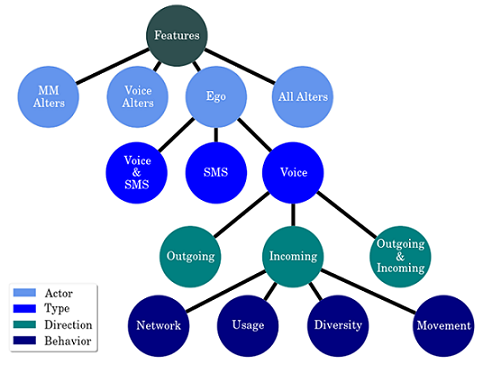}
		\caption{Tree-based feature classification \label{fig:tree}}
	\end{minipage}
	%	\caption{Feature engineering. \label{fig:features}}
\end{figure*}

The most common approach to constructing interpretable metrics (``features") from the phone data is to hand-craft a small number of features that correspond to some intuition of the researcher. For instance, \cite{dong_who_2014} focus on 5 topological properties of the static social network. In contrast, we employ a deterministic finite automaton (DFA), to formalize the feature generation process \cite{mcculloch_logical_1943,blumenstock_predicting_2015}. We use the DFA to specify a set of legal operations that can be recursively applied to raw transactional data in order to produce valid features. The DFA we use to generate features from CDR is shown in Figure \ref{fig:states} (see also Algorithm \ref{algorithm}). To interpret the set of features produced by the DFA, we map each feature onto the tree structure as shown in Figure \ref{fig:tree}, where each branch of the tree partitions the data into features that capture different behavioral characteristics of the subscriber.

\newcommand{\Let}[2]{\State #1 $\gets$ #2}

\setlength{\textfloatsep}{45pt}% Remove \textfloatsep

\begin{algorithm}[!ht]

	\SetAlgoLined %\SetAlgoNoLine
	\caption{Feature Generation Algorithm\label{algorithm}}
\BlankLine
	\KwData{cdr (Call Detail Records of all users)}
	\KwData{opmap (Dictionary of possible operations)}
%	$cdrtypes \leftarrow$ [Voice, SMS, Voice and SMS]\\
%	$direction \leftarrow$ [In, Out , In plus Out]\\
%	$featuresArray \leftarrow$ []

	\KwResult{Features}
	\hrulefill

	Step 1: Perform reduce by grouping only on ego
	\ForEach {$type$, $dir1$ in {$cdrtypes$,$direction$}}{
		
			$filteredCDR \leftarrow$ $cdr$.filter($type$,$dir1$)\\
			\ForEach {$field$ in $cdr$}{
				$groupeddata \leftarrow$ $filteredCDR$.map([$ego$] +[combinations($field$])) \\
				\ForEach {$op$ in $opmap$[$field$]}{
					$reduceddata \leftarrow$ $groupeddata$.reduce($op$)
					insert $reduceddata$ in $featuresArray$\\
					$reducedata2 \leftarrow$ $reduceddata$.map($ego$, $alters$).reduce($op$)
					insert $reduceddata2$ in $featuresArray$\\
				}	
				
			}	
	}

\end{algorithm}

%%%%%%%%%%%%%%%%%%%%%%%%%%%%%%%%
%% METHODS
%%%%%%%%%%%%%%%%%%%%%%%%%%%%%%%%%
%\section{Modeling}

%\begin{figure*}[ht!]
%	\centering
%	\begin{minipage}[t]{0.48\linewidth}
%		\includegraphics[width=3.3in]{./figs/UserCounts.png}%
%		\caption{Distribution of user types by country\label{fig:counts}}%
%	\end{minipage}
%	\quad  
%	\begin{minipage}[t]{0.48\linewidth}
%		\includegraphics[width=3.3in]{./figs/ghana_calls.png}%
%		\caption{Distribution of calls per subscriber, Ghana\label{fig:ghana_calls}}%
%	\end{minipage}
%\end{figure*}

%\input{methods.tex}

%%%%%%%%%%%%%%%%%%%%%%%%%%%%%%%%
% RESULTS
%%%%%%%%%%%%%%%%%%%%%%%%%%%%%%%%

\section{Results}
\label{sec:results}
\label{sec:methods}

%The DFA-based process of feature engineering described above generates thousands of features that quantify patterns of mobile phone use. Using with these features, our goals are 
Using the features generated by DFA \ref{sec:features} our next steps were to (a) build a predictive model that can be used to identify likely adopters, (b) use these features to understand the determinants of Mobile Money use, and (c) determine the extent to which models and features from one context can generalize to another.

\subsection{Predicting Mobile Money Use}
\begin{figure*}[htb]
	\centering
	\begin{minipage}[t]{0.48\linewidth}
		\includegraphics[width=3in]{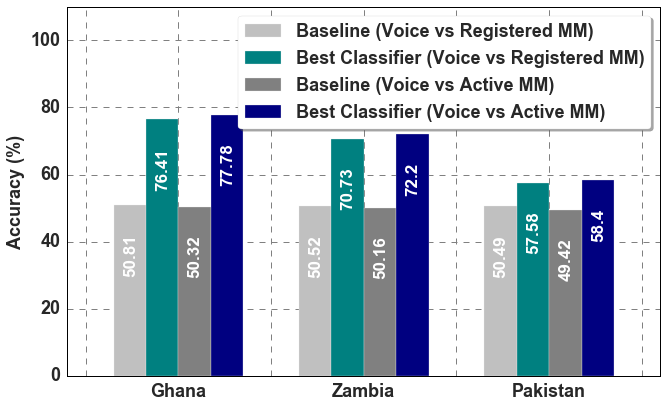}%
	\end{minipage}
	%	\quad 
	\begin{minipage}[t]{0.48\linewidth}
		\includegraphics[width=3in]{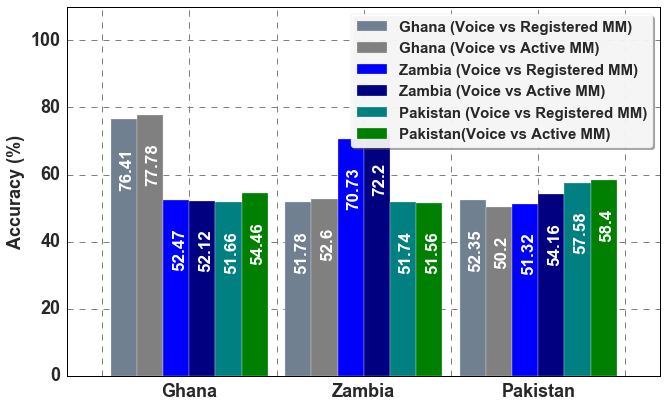}%
	\end{minipage}
	\caption{Model Accuracy. Left figure shows accuracy in identifying Mobile Money users within each country; Right figure shows Accuracy when model is trained in one country and evaluated in another\label{fig:accuracy}.}
\end{figure*}

We use a variety of supervised learning algorithms to attempt two classification tasks: Voice Only vs. Registered User, and Voice Only vs. Active User. For this task, we drew a stratified random sample of 10,000 subscribers from each of the three user types from each of the three countries. Here, we focus on the results from gradient boosting \cite{friedman2001greedy}, which marginally outperformed other common classifiers including logistic regression, svm, and so forth.
As shown in the Figure \ref{fig:accuracy}, the gradient boosting with DFA based features significantly outperforms the naive baseline, and we achieve marginally better success in identifying Active Mobile Money users than Registered Mobile Money users in all of the countries. We will return to the interpretation of these results in Section \ref{sec:discussion}.

\subsection{Determinants of Mobile Money Adoption}

To investigate the prominant denominators of Mobile Money adoption, we calculate the \textit{AUC} from a bivariate logistic regression of the response variable on each individual feature. We also calculate the \textit{normalized feature importance (NFI)} as the {relative} significance of each feature in the final gradient boosting classifier operating on all of the features.

\begin{figure*}[htb!]
	\centering
	\begin{minipage}[t]{0.48\linewidth}
		\includegraphics[width=.92\textwidth]{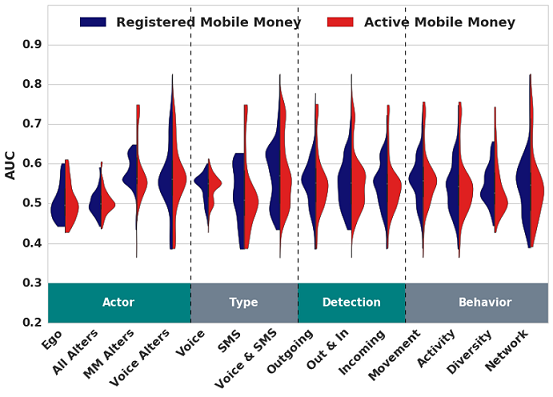}
	\end{minipage}
	%	\quad 
	\begin{minipage}[t]{0.48\linewidth}
		\includegraphics[width=.92\textwidth]{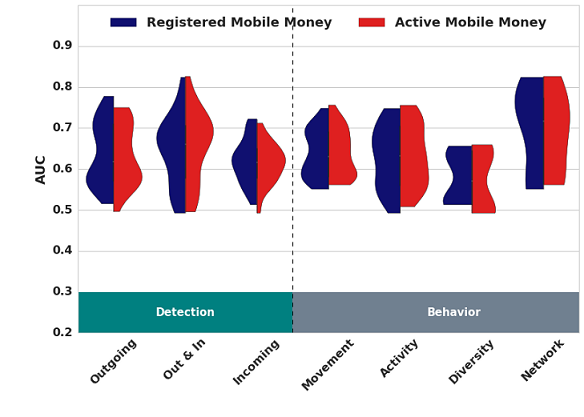}
	\end{minipage}
	\caption{Distribution of AUC values for each feature category. Left figure shows all features in Ghana; Right figure shows the subset of features in Ghana where Actor=`Voice Alters' and Type=`All'\label{fig:ghana_vio}}
\end{figure*}

The distribution of unconditional AUC values for all the features generated through the methods explained in section \ref{sec:features} is shown in Figure \ref{fig:ghana_vio} (left panel), using Ghana as a test case.  Each violin plot shows the distribution of AUC values for all features of a given type - such as all ``ego'' features, or all ``movement'' features. While a large number of features have AUC values near 0.5,  there are a small number of highly predictive features with AUC$\geq$0.75. The right panel of Figure \ref{fig:ghana_vio} shows the distribution of AUC values for the subset of features related to the subscriber's first degree network (Actor=``Voice Alters'' and Type=``All''), a subset that is generally more predictive of Mobile Money use in Ghana. Here, the range of AUC values is significantly higher than in the full set of features, and some sub-classes such as ``Network'', which capture the structure of the subscriber's first degree network, have uniformly high predictive power.%

\begin{figure*}[htb]
	\centering
	\begin{minipage}[t]{0.32\linewidth}
		\includegraphics[width=.98\textwidth]{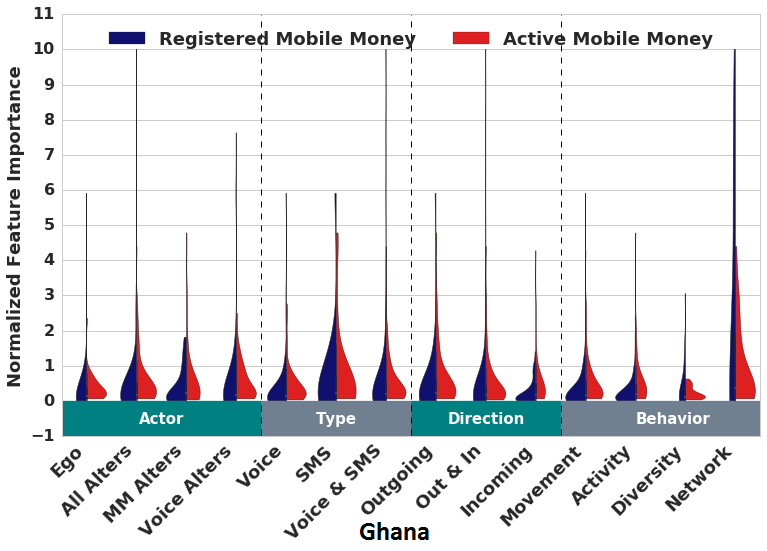}
	\end{minipage}
	%	\quad 
	\begin{minipage}[t]{0.32\linewidth}
		\includegraphics[width=.98\textwidth]{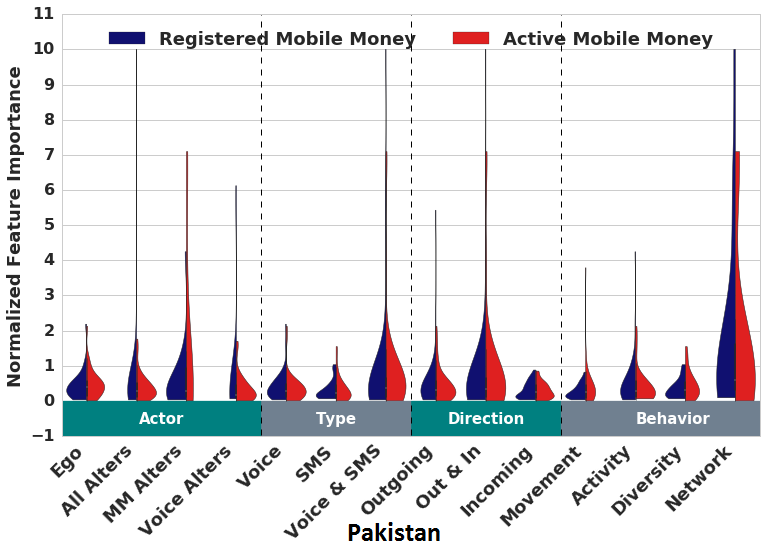}
	\end{minipage}
	%	\quad 
	\begin{minipage}[t]{0.32\linewidth}
		\includegraphics[width=.98\textwidth]{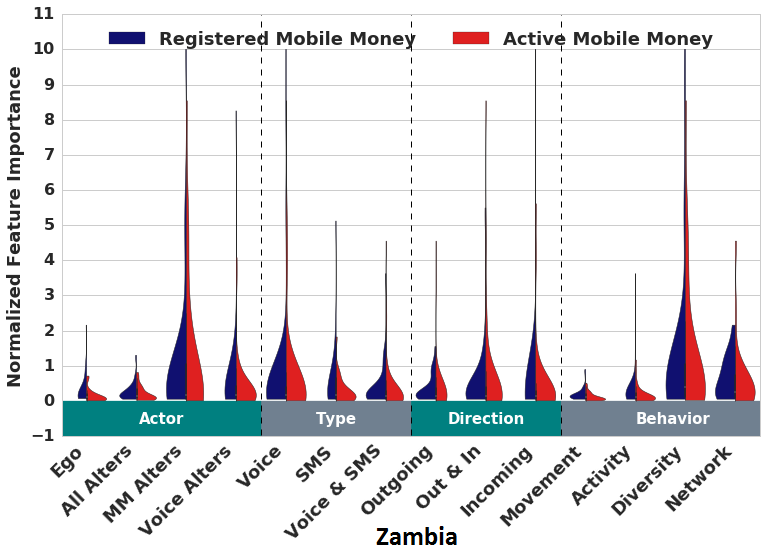}
	\end{minipage}
	\caption{Normalized feature importance\label{fig:nfi}}
\end{figure*}

Figure \ref{fig:nfi} shows the NFI distribution obtained through gradient boosting model operating on all of the features. Just like unconditional ranking, each class of features in this case also contains a mass of features with low predictive power, but perhaps most striking in Figure \ref{fig:nfi} are the differences between countries in the relative importance of each class of features. 

\subsection{Learning Across Cultures?}

Our standardized methods and analysis performed across all three countries. This makes it possible to answer a question that has been elusive in prior studies of the adoption of new technologies in developing countries: \textit{Do the behavioral determinants of adoption identified in one context generalize to another?} Based on the analysis we have performed, our short answer to this question appears to be, ``No."

The right panel of Figure \ref{fig:accuracy} shows the performance of a classifier trained in one country and evaluated in another. The first set of six bars shows that the classifiers trained in Ghana perform well in Ghana (the first two grey bars), replicating the results in Figure \ref{fig:accuracy}. However, that same Ghana model does quite poorly when evaluated in Zambia (the next two blue bars) or Pakistan (the final two green bars). While it is almost certain that a more sophisticated approach to transductive transfer learning would perform better \cite{pan_domain_2011}, the naive application of a model out of context is quite ineffective. 

%%%%%%%%%%%%%%%%%%%%%%%%%%%%%%%%
% DISCUSSION
%%%%%%%%%%%%%%%%%%%%%%%%%%%%%%%%
\section{Discussion and Conclusions}
\label{sec:discussion}
\label{sec:conclusion}

Taken in the broader context of research into the determinants of Mobile Money adoption, the preceding results uncover several unexpected patterns.It is not surprising that classifiers using CDR-data can predict Mobile Money use. However, we were surprised to find that the supervised model was only marginally better able to identify Active Mobile Money users than Registered Mobile Money users. Since true financial inclusion requires active use, this remains an important topic for future work.%

Also interesting are the differences in performance of the same modeling approach applied in different contexts. Most striking here is the relatively poor performance in Pakistan, where the 18\% improvement over the baseline is dwarfed by the 55\% improvement over the baseline achieved in Ghana. We believe this may also in part be an artifact of the ``one size fits all'' approach we have taken to standardizing definitions and methods across countries. In particular, there is one type of Mobile Money transaction that is extremely common in Pakistan, which allows a subscriber to add prepaid phone credit to her phone account using Mobile Money. Anecdotally, it is common practice in Pakistan for the retailers of phone credit to perform this Mobile Money transaction on behalf of the subscriber. This potential source of bias highlights the brittle nature of the cross-country analysis, which in its current form does not allow for country-specific adaptation.

Most importantly, our results suggest that across different countries in the developing world, no single set of behavioral features is likely to consistently predict Mobile Money adoption and use. For policymakers interested in promoting the growth of mobile-based financial services, this indicates that the details of the local context will be critical in achieving sustained adoption and use of financial services for the poor.

%%%%%%%%%%%%%%%%%%%%%%%%%%%%%%%%
% CONCLUSION
%%%%%%%%%%%%%%%%%%%%%%%%%%%%%%%%
% \section{Conclusion}
%\input{conclusion.tex}

\clearpage
\bibliography{mm_icml_references}
\bibliographystyle{icml2016}

\end{document}